\documentclass[twoside,a4paper]{article}

\usepackage[textwidth=15cm]{geometry}

\usepackage{fancyhdr}

\usepackage{graphicx}
\usepackage{multirow}
\RequirePackage{booktabs} 
\usepackage{arydshln}

\usepackage{hyperref}

\usepackage{xspace}
\usepackage{xcolor}
\usepackage{amsmath,amsfonts,amssymb}
\usepackage{dsfont}
\usepackage{bm}

\newcommand{\VCN}{\textsf{VCNet}\xspace}

\title{VCNet: A self-explaining model for \\realistic counterfactual generation}

\author{
Victor Guyomard$^{1,2}$, Françoise Fessant$^{1}$, Thomas Guyet$^{3}$\\ Tassadit Bouadi$^{2}$, Alexandre Termier$^{2}$
}

\date{1. Orange Labs, Lannion, France\\
\href{mailto:victor.guyomard@orange.com}{victor.guyomard@orange.com}\\
2. Univ Rennes, Inria, CNRS, IRISA, Rennes, France \\
3. Inria, Centre de Lyon, France}

\begin{document}

\maketitle

\noindent\makebox[\textwidth][c]{
    \begin{minipage}{.75\textwidth}
        Counterfactual explanation is a common class of methods to make local explanations of machine learning decisions. 
For a given instance, these methods aim to find the smallest modification of feature values that changes the predicted decision made by a machine learning model. 
One of the challenges of counterfactual explanation is the efficient generation of realistic counterfactuals. 
To address this challenge, we propose \VCN\ -- Variational Counter Net -- a model architecture that combines a predictor and a counterfactual generator that are jointly trained, for regression or classification tasks. 
\VCN is able to both generate predictions, and to generate counterfactual explanations without having to solve another minimisation problem. 
Our contribution is the generation of counterfactuals that are close to the distribution of the predicted class. 
This is done by learning a variational autoencoder conditionally to the output of the predictor in a join-training fashion. 
We present an empirical evaluation on tabular datasets and across several interpretability metrics. The results are competitive with the state-of-the-art method.
    \end{minipage}}

\pagestyle{fancy}

\section{Introduction} 
Improvements of machine learning techniques for decision systems has led to the rise of applications in various domains such as healthcare, credit or justice. 
The eventual sensitivity of such domains, as well as the black-box nature of the algorithms, has motivated the need for methods that explain why some prediction was made. 
For example, if a person's loan is rejected as a result of a model decision, the bank must be able to explain why. 
In such a context, it might be interesting to provide an explanation of what that person should change to influence the model's decision. 
As suggested by Wachter et al.~\cite{Wachter2017CounterfactualEW}, one way to build this type of explanation is through the use of counterfactual explanations. 
A counterfactual is defined as the smallest modification of feature values that changes the prediction of a model to a given output. 
In addition, the explanation also provides important feedback to the user. 
In the context of a denied credit, a counterfactual is a close individual for whom his credit is accepted and the feature modifications suggested by a counterfactual acts as recourse for the user. 
For privacy reason or simply because there is no similar individual with an opposite decision, we aim to generate synthetic individuals as counterfactuals.
In order to provide a meaningful recourse, the counterfactual is expected to be realistic, i.e. close to the existing examples and respecting the observed correlation among features. Furthermore, in order to be representative of its predicted class, it is interesting to obtain a counterfactual close to the existing examples but relative to the counterfactual class. 
A counterfactual instance is usually found by iteratively perturbing the input characteristics of the original example until the desired prediction is achieved, which is like minimizing a loss function using an optimization algorithm. Many methods proceed in this way, but differ in their definition of the loss function and optimization method \cite{molnar2022}. These approaches appear to be computationally expensive. Indeed, for each instance to explain, a new optimisation problem has to be solved. 
Most of the counterfactual methods apply to already trained decision models and treat them as black boxes in the post-hoc paradigm. However, dissociating the prediction of the model from its explanation can lead to an explanation of poor quality~\cite{2019rudin}.

Self-explaining models which incorporate an explanation
generation module into their architecture, such that they provide explanations for their own predictions, can be an alternative to the previous approaches. In general, the predictor and explanation generator are trained jointly, hence the presence of the explanation generator is influencing the training of the predictor~\cite{2020Elton}. In this spirit, Guo et al.~\cite{guo2021counternet} propose CounterNet, a neural network based architecture for the prediction and counterfactual generation along with a novel variant of back propagation to ensure the stabilization of the training process. Compared to a post-hoc approach, they are able to produce counterfactuals with higher validity. A counterfactual is said to be valid if it succeeds in reaching a different prediction.    
A limitation of the CounterNet approach is that counterfactuals it generates may lack realism w.r.t. the data points of the class where they are positioned. 
The proposed approach, \VCN, tackles this issue: similarly to CounterNet, it combines a predictor and a counterfactual generator that are jointly trained. The difference lies in the counterfactual generator based on conditional variational autoencoder (cVAE) whose latent space can be controlled and tweaked to generate realistic counterfactuals. Our approach is inspired from John et al. work about learning disentangled latent spaces in the context of text style transfer~\cite{johnetal2019disentangled}.

Our main contribution is the proposal of a cVAE for counterfactual generation in order to generate realistic counterfactuals. 
Our second contribution is a self-explainable architecture of a classifier that embeds a cVAE, used as a counterfactual generator. In this architecture, both the model and the cVAE, are jointly trained with an efficient single back-propagation procedure. After recalling the properties of the variational autoencoder models (Section~\ref{sec:backgrounds}) that interest us in the context of this paper, we describe our proposal (Section~\ref{sec:vcnet}). Then we present extensive experimental studies on synthetic and real data (Section~\ref{sec:expes}). We compare the quality of the counterfactuals produced with those of CounterNet on different datasets through state-of-the-art metrics. The focus is on tabular data but we also show that our architecture is interesting on images.

\section{Related Work}\label{sec:soa}
Our work is concerned with the search for counterfactual explanation that is usually found by iteratively perturbing the input features of the instance of interest until the desired prediction is reached. 
In practice, the search of counterfactuals is usually posed as an optimization problem. It consists of minimizing an objective function which encodes desirable properties of the counterfactual instance with respect to the perturbations. Wachter et al.~\cite{Wachter2017CounterfactualEW} propose the generation of counterfactual instances by minimizing the distance between the instance to be explained and the counterfactual while pushing the new prediction toward the desired class. Other algorithms optimize other aspects with additional terms in the objective function such as actionability~\cite{UstunFAT19}, diversity~\cite{MothilalFAT20,RusselFAT19},  realism~\cite{VanLooverenECML21}. 

All aforementioned techniques search for counterfactual example by solving a separate optimization problem for each instance to be explained. This optimization problem is computationally intensive, making it impractical for large numbers of instances. To address this issue, several frameworks based on  generative models have been proposed. A generative counterfactual
model is trained to predict the counterfactual perturbations or instances directly. Many of these frameworks use the latent space of variational autoencoder models to generate counterfactuals with linear interpolation~\cite{barr2021counterfactual}, latent feature selection~\cite{downs2020cruds}, perturbation~\cite{PawelczykWWW20} or incorporation of the target class in the latent space~\cite{Nangi2021CounterfactualsTC}). 

All these counterfactual generation techniques are post-hoc as they assume that the explaining task is done after the decision task with a fixed black-box model. In this post-hoc paradigm, the counterfactual search process is completely uninformed from the decision process. Post-hoc explanations may be the only option for already-trained models. 

Another approach is to design models that optimize both the decision and an explanation
of that decision during the learning process \cite{NEURIPS2018_3e9f0fc9}. In the context of counterfactual explanations, such a strategy has been recently proposed by Guo et al.~\cite{guo2021counternet} with CounterNet, a framework in which prediction and explanation are jointly learned. The optimization of the counterfactual example generation only once together with the predictive model allows a better alignment between predictions and counterfactual explanations. This leads to explanations of better quality and significantly reduces the generation process time. Our architecture is inspired by theirs but we have chosen to use a conditional variational autoencoder as a generative model of counterfactuals allowing us to exploit text style-transfer techniques~\cite{johnetal2019disentangled,Nangi2021CounterfactualsTC}.
%

\section{Backgrounds}\label{sec:backgrounds}
In this section, we introduce some notations, problematic and backgrounds necessary for the presentation of the \VCN architecture in the next section.

Let $\mathcal{X}\subseteq \mathbb{R}^p$ represents the $p$-dimensional feature space and $l \geq 2$ a number of classes. A training dataset, denoted $D = \{(\bm{x_i} , y_i )\}_{i=1}^n$, is such that $\bm{x_i} \in \mathcal{X}$ and $y_i \in \{1, \dots, l\}$ for each $i \in \{1, \dots, n\}$. 
For a new example $\bm{x}$, the prediction consists in deciding to which class $\hat{y}$ the example $\bm{x}$ belongs. For more generality, we consider probabilistic prediction: the prediction is a probability vector, denoted $\bm{\hat{p}}\in[0,1]^p$. Then, the predicted class is the most likely class according to $\bm{\hat{p}}$.
The counterfactual generation yields an example $\bm{x'}$ which is close to $\bm{x}$ and whose prediction $\hat{y}'$ is different from $\hat{y}$ in case the counterfactual is valid.

Our problem is both to learn from ${D}$ an accurate predictor, denoted $f$, and a generator of counterfactual, denoted $g$.

Now that we have presented our problem, we introduce the notion of VAE~\cite{VAE_original} and cVAE~\cite{CVAE_original} which our proposal relies on.

\subsection{Variational Autoencoder (VAE)} 
A variational autoencoder is a generative model where a latent parameterized distribution is learned. If samples are drawn in the latent space according to this distribution, the decoded samples are expected to be distributed according to the training data distribution (an approximate distribution of the training data distribution is learned)~\cite{VAE_original}. 
Formally, let $\bm{z}$ be a latent vector (drawn from the latent distribution) and  $\bm{x}$ be an example, we denote by
$q_{\phi}\left(\bm{z} \mid \bm{x}\right)$ the encoder distribution and by $p_{\theta}\left(\bm{x} \mid \bm{z}\right)$ the decoder distribution. 
Training a VAE is finding the parameters $\theta$ and $\phi$ that minimize the following objective function, i.e. the opposite of the Evidence Lower Bound (ELBO): 
\begin{equation}\label{eq:elbow}
\mathcal{L}_{\text{VAE}}(\theta,\phi)=-\mathbb{E}_{q_{\phi}\left(\bm{z} \mid \bm{x}\right)}\left[\log (p_{\theta}\left(\bm{x} \mid \bm{z}\right))\right]+\mathrm{D}_{K L}\left[q_{\phi}\left(\bm{z} \mid \bm{x}\right) \Big\Vert\, p(\bm{z})\right]
\end{equation}
Generally, distributions are chosen to be Gaussian, meaning that $q_{\phi}\left(\bm{z} \mid \bm{x}\right)\sim \mathcal{N}\left(\mu_{\phi}, \Sigma_{\phi}\right)$ and $p_{\theta}\left(\bm{x} \mid \bm{z}\right)\sim \mathcal{N}\left(\bm{x}\mid\mu_{\theta}, \Sigma_{\theta}\right)$ and distribution parameters are estimated thanks to back-propagation. 
The first term of Eq.~\ref{eq:elbow} encourages reconstructing $\bm{x}$ at the output of the decoder ($\hat{\bm{x}}$). 
The second term encourages the regularization of the latent space by pushing $q_{\phi}\left(\bm{z}\mid \bm{x}\right)$ to a Gaussian prior $p\sim\mathcal{N}(0, I)$.

\subsection{Conditional Variational Autoencoder (cVAE)}\label{sec:cvae}
A conditional variational autoencoder is a VAE where distributions are conditioned on a given variable $c$ \cite{CVAE_original}. 
The architecture is the same as a standard VAE but $c$ is given as input of the encoder and also as input of the decoder. 
Then the objective function of Eq.~\ref{eq:elbow} can be rewritten as: 
\begin{multline} \label{conditional_elbow}
\mathcal{L}_{\text{cVAE}}\left( \theta ,\phi \right) = -\mathbb{E}_{q_{\phi}\left(\bm{z} \mid \bm{x},c\right)}\left[\log (p_{\theta}\left(\bm{x} \mid \bm{z},c\right))\right]+\mathrm{D}_{K L}\left[q_{\phi}\left(\bm{z} \mid \bm{x},c\right) \Big\Vert\, p(\bm{z})\right]
\end{multline}
The encoder distribution becomes $q_{\phi}\left(\bm{z} \mid \bm{x},c\right)$ and is pushed to the Gaussian prior $p\sim\mathcal{N}(\bm{z} \mid \bm{0}, \bm{I})$ by the regularization term regardless of the value of $c$. 
The decoder reconstructs $\bm{x}$ from the concatenation of $\bm{z}$ with $c$.

The initial objective of conditional variational autoencoder is to enrich the expressiveness of the model in supervised settings. 
In this article, we use a property of the cVAE to disentangle the class specification from the content of the data \cite{kingma2014semi}.\footnote{For Kingma et al~\cite{kingma2014semi}, what we call the ``content'' in this paper is denoted the ``style''. It refers to the writing style of digits in MNIST-like datasets.} 
Intuitively, the latent variable $\bm{z}$ does not need to model the example category, then it can focus on modeling the content of the examples, which is shared by all the categories. 
To illustrate the effective disentanglement of category and content, Kingma et al. show that the decoder $p_\theta\left(\bm{x} \mid \bm{z},c\right)$ can be used to generate images of the ten digits with the same shared content (let say the handwriting) by changing the class $c$ and keeping the same random values for $\bm{z}$. 
The same property has been applied to text style transfer~\cite{johnetal2019disentangled}. In this context, the style is the category, and the content is the wording. 
For tabular data, the notion of ``content'' and ``style'' can be illustrated in the context of the loan decision. The ``style'' characterizes the category of people loan (\textit{accepted} or \textit{rejected}) and the ``content'' characterizes the other features. More specifically, the later models correlations between variables that are independent from the loan decisions.

In our proposal, we exploit the modeling properties of a cVAE to generate counterfactuals. Considering that the cVAE disentangles the category and the content, the decoder of a cVAE can be tweaked in a flexible way. For $\bm{z}$ the encoding of an example $\bm{x}$ of class $c$, $p_\theta\left(\bm{x} \mid \bm{z},c'\right)$ generates examples of a category $c'\neq c$. 
In addition, $(\bm{z},c')$ is likely the element of class $c$ that is the closest to $(\bm{z},c)$ in the latent space. As this space is regularized, $p_\theta\left(\bm{x} \mid \bm{z},c'\right)$ generates examples that are similar to $\bm{x}$, but belonging to a different class. This fits exactly the expectations of counterfactuals and will be assessed in Section~\ref{sec:results:synth}.

\section{A Join Training Model}\label{sec:vcnet}
\VCN is a self-explainable \footnote{By Self-explainable model here we mean that the predictor is constrained by the counterfactual generator during training but the explanation is not directly used to produce model output as in~\cite{NEURIPS2018_3e9f0fc9}.}  model through counterfactual generation. 
Inspired by CounterNet~\cite{guo2021counternet}, the \VCN model is made of a predictor, $f(\cdot)$, and a counterfactual generator, $g(\cdot,\cdot)$, that are jointly trained. 
In the inference phase, each part can be used on demand: on the one side, to get the prediction $f(\bm{x})$ for some new example $\bm{x}$ and, on the other side, to generate its counterfactual $g(\bm{x},c)$ for another class $c$. 
\VCN can be used as a self-explainable model and generates $\left(f\left(\bm{x}\right), g\left(\bm{x},c\right)\right)$, i.e. the couple of the prediction and its counterfactual.

The trick of \VCN is to not train a counterfactual generator, but a supervised autoencoder, i.e. a cVAE. The cVAE is trained as an autoencoder and used as a counterfactual generator in inference. 

We start by presenting the architecture of our network, then we detail the training problem by defining the loss which is optimized and finally, we detail how the trained model is used to generate counterfactuals.

\subsection{\VCN Architecture}
\begin{figure}[t!]
    \centering
    \includegraphics[width=.98\textwidth]{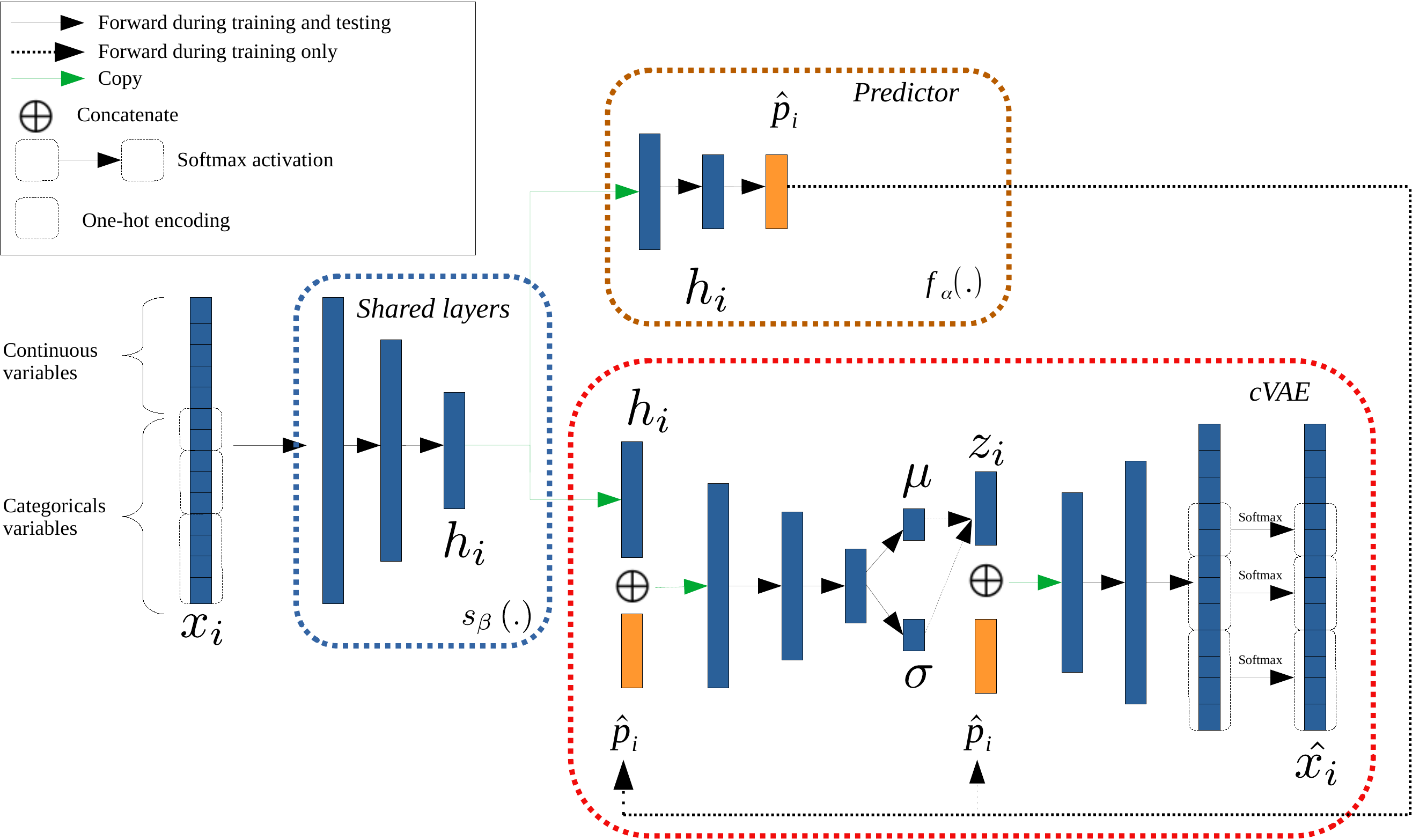}
    \caption{\VCN architecture is composed of three blocks: Shared layers that transform the input into a latent representation (blue square), a predictor that outputs the prediction (brown square), and a conditional variational autoencoder that acts as a counterfactual generator during testing (red square).}
    \label{fig:network_archi}
\end{figure}

\VCN is a neural network architecture. 
Figure~\ref{fig:network_archi} illustrates this architecture made of three main blocks:
\begin{enumerate}
    \item Shared layers, $s_\beta$, that transform the input into a dense latent representation. 
    We use fully connected layers with ELU activation functions.
    
    \item A predictor network $f_\alpha$ that takes the shared latent representation  and returns a probability vector corresponding to the probabilistic prediction. 
    We use fully connected layers with ELU activation functions.
    
    \item A conditional variational autoencoder that takes as input the shared latent representation and also the probability vector given by the predictor. The cVAE reconstructs examples and integrates additional layers to handle categorical variables (see details below).
\end{enumerate}
During training, an example $\bm{x}_{i}$ is passed through the shared layers to generate a dense latent representation $\bm{h}_{i}=s_\beta(\bm{x}_{i})$. This representation is then shared with both the predictor network and the autoencoder. 
On the one hand, $\bm{h}_{i}$ is passed through the predictor $f_\alpha$ in order to obtain the probability vector $\hat{\bm{p}_{i}}$. Then, the prediction of an example $\bm{x}_{i}$ is obtained by the function $f_{\alpha,\beta}(\bm{x}_i)=f_{\alpha}\circ s_\beta(\bm{x}_{i})$.
On the other hand, $\bm{h}_{i}$ is passed through the cVAE. 
It is first concatenated with $\hat{\bm{p}}_{i}$ and then is passed through the encoder of the cVAE and samples a latent vector $\bm{z}_{i}$. 
This latent vector concatenated with $\hat{\bm{p}_{i}}$ is passed through the decoder, so as to obtain a reconstructed example $\hat{\bm {x}}_{i}$.

It can be noticed that the cVAE slightly differs from the original cVAE \cite{CVAE_original}. Indeed, formally, the encoder of an end-to-end cVAE includes the shared layer, $s_\beta$. In our architecture, the condition is introduced at an intermediary latent representation of the examples. 
The idea behind this architecture is to enforce the dense latent representation to be as independent of the class as possible. 

In addition, we adopt the same preprocessing as Guo et al.~\cite{guo2021counternet} to handle categorical variables. 
At the input of the network, each categorical variable is encoded with a one-hot. 
At the output of the cVAE, we add a softmax activation function for each one-hot categorical variable in order to obtain a one-hot encoding format by taking the \textit{argmax}.
Finally, continuous variables are scaled to have all variables in $\left[0,1\right]$.

\subsection{Loss Function and Training Procedure} 
The objective of the training is to jointly learn the predictor and the cVAE. Then, the loss to minimize is made of two terms.

The first term is derived from the loss of a cVAE defined in Eq.~\ref{conditional_elbow}. In our case, the cVAE is conditioned by the probability vector obtained at the output of the predictor. 
Then we can rewrite the objective function as: 
\begin{multline} \label{eq:our_elbow}
\mathcal{L}_{cVAE}\left( \theta ,\phi ,\beta; \bm{x}_i \right) = -\lambda_{3}\mathbb{E}_{q_{\phi}\left(\bm{z}_{i} \mid s_{\beta}\left(\bm{x}_{i}\right),\hat{\bm{p}}_{i}\right)}\left[\log (p_{\theta}\left(\bm{x}_{i} \mid \bm{z}_{i},\hat{\bm{p}}_{i}\right))\right]\\+\lambda_{1}\mathrm{D}_{K L}\left[q_{\phi}\left(\bm{z}_{i} \mid s_{\beta}\left(\bm{x}_{i}\right),\hat{\bm{p}}_{i}\right) \Big\Vert\,  p(\bm{z}_{i})\right]
\end{multline} 
$\lambda_{1}$ and $\lambda_{3}$ are weights to control the impact of each term during the training phase. 

The second term enables us to learn the predictor. We use cross-entropy as classification loss between the output of the predictor $\hat{\bm{p}}_{i} = f_\alpha\circ s_\beta(\bm{x}_i)$  and the true label $y_{i}$:
\begin{equation}\label{eq:predictor}
\mathcal{L}_{pred}\left( \alpha,\beta;\bm{x}_i,y_i\right) = \sum_{k=1}^{l}-\mathds{1}_{\left[y_{i}=k\right]} \log \left(\left[f_\alpha\circ s_\beta(\bm{x}_i)\right]_{k}\right)
\end{equation}
where $\left[f_\alpha\circ s_\beta(\bm{x}_i)\right]_{k}$ denotes the predicted probability that $\bm{x}_i$ belongs to the $k$-th class.

Then, the loss function on the training set ($D$) is a weighted sum of the losses from Eq.~\ref{eq:predictor} and Eq.~\ref{eq:our_elbow} as follows:
\[
\mathcal{L}\left( \theta ,\alpha,\beta,\phi; D\right) = \sum_{i=1}^{n} \mathcal{L}_{cVAE}\left( \theta ,\phi ,\beta;\bm{x}_i \right) + \lambda_{2} \frac{1}{n} \sum_{i=1}^{n} \mathcal{L}_{pred}\left( \alpha,\beta;\bm{x}_i,y_i \right)
\]

As mentioned at the beginning of this section, it is worth noticing that our problem is not to learn to generate counterfactuals. Then, contrary to CounterNet that has divergent objectives to optimize, the minimization of $\mathcal{L}$ is a simple optimization problem solved by back-propagation. 

Note that $\lambda_{1}$,$\lambda_{2}$,$\lambda_{3}$ are hyperparameters to tune for training.

\subsection{Counterfactual Generation} 
Since our model does not directly produce counterfactuals, some modifications are needed for inference. 
An example $\bm{x}_{i}$ is passed through the prediction network to obtain both its predicted probability vector ($\hat{\bm{p}}_{i}$) and its dense vector representation ($\bm{h}_i$). 
This dense vector representation ($\bm{h}_i$) and the predicted probability vector ($\hat{\bm{p}}_{i}$) are given to the encoder of the cVAE to produce a latent vector $\bm{z}_i$. 
Then, the decoder of the cVAE plays the role of a counterfactual generator.
Because we want to generate an example with a different predicted class we need a probability vector $\bm{p}_{c}$ such that the class with maximum probability is different from the one of the prediction, i.e.  $\arg\max\left(\hat{\bm{p}}_{i}\right) \neq \arg\max\left( \bm{p}_{c}\right)$.
In a binary-classification problem setup, we decided to use a one-hot vector where the probability is 0 for the predicted class of $x_{i}$ and 1 for the opposite class. 
The reason for this choice is that we want to generate counterfactuals for which the confidence in the predicted class is the highest for the predictor. 
In the case of a multi-class classification problem, we propose to select the class with the second-highest probability in $\hat{\bm{p}}_{i}$ and to switch the values with the predicted class. For example, if we obtain a probability vector $\hat{\bm{p}}_{i} = \left[0.6,0.3,0.1\right]$ then $\bm{p}_{c} = \left[0.3,0.6,0.1\right]$.
An alternative solution would be to let the user select the class for which he/she is interested in having a counterfactual. 
     
This vector $\bm{p}_{c}$ and the dense latent representation $\bm{z}_i$ are passed through the cVAE decoder in order to infer a new predicted class to obtain a counterfactual \footnote{Note that the quality of the generated counterfactual depends on the quality of the learned latent space.} $\bm{x}_{c}$. 
As explained in Section \ref{sec:cvae}, the intuition is to benefit from the disentanglement of the latent space of a cVAE: 
$\bm{z}_i$ contains non-class-specific information about $\bm{x}_{i}$ and 
$\bm{p}_{c}$ encodes information for the desired class. As such, the decoder generates a new example $\bm{x}_{c}$ that is similar to $\bm{x}_{i}$ and that belongs to a different class.

\section{Experiments and Results}\label{sec:expes}
Our experiments are organized in four steps. 
Our main objective is to show that \VCN generates counterfactuals that are both valid (counterfactuals actually belong to another class) and realistic (counterfactuals are close to examples of the same class). In the first set of experiments, we present results of a cVAE on a synthetic dataset to confirm the actual disentanglement of the content and the class. These experiments also aim at providing some intuition about the reason why \VCN works. 
In the second set of experiments, we compare the results of \VCN with CounterNet, the state-of-the-art algorithm for self-explainable counterfactual generation. 
The reader interested in the results of more counterfactual generation systems may refer to the original article of Guo et al.~\cite{guo2021counternet}. Our experiments use the same datasets and data preprocessing. 
In the third experiment, we evaluate the impact of join training on the quality of counterfactuals. We propose a post-hoc version of our framework and compare the results obtained with the jointly trained \VCN.
Finally, we present some qualitative experiments on the MNIST dataset. We choose to present experiments on MNIST firstly because it has been widely used in the field of counterfactual generation and, secondly, because it illustrates that \VCN may be applied on different types of data (tabular, images, time series, ...).

The hyperparameters and architectures of the models used in these experiments are detailed in Supplementary material. The code to reproduce the
results of this section is also provided in supplementary material.
\subsection{cVAE for counterfactual generation}\label{sec:results:synth}
\begin{figure}[tbp]
\centering
\includegraphics[width=.85\textwidth]{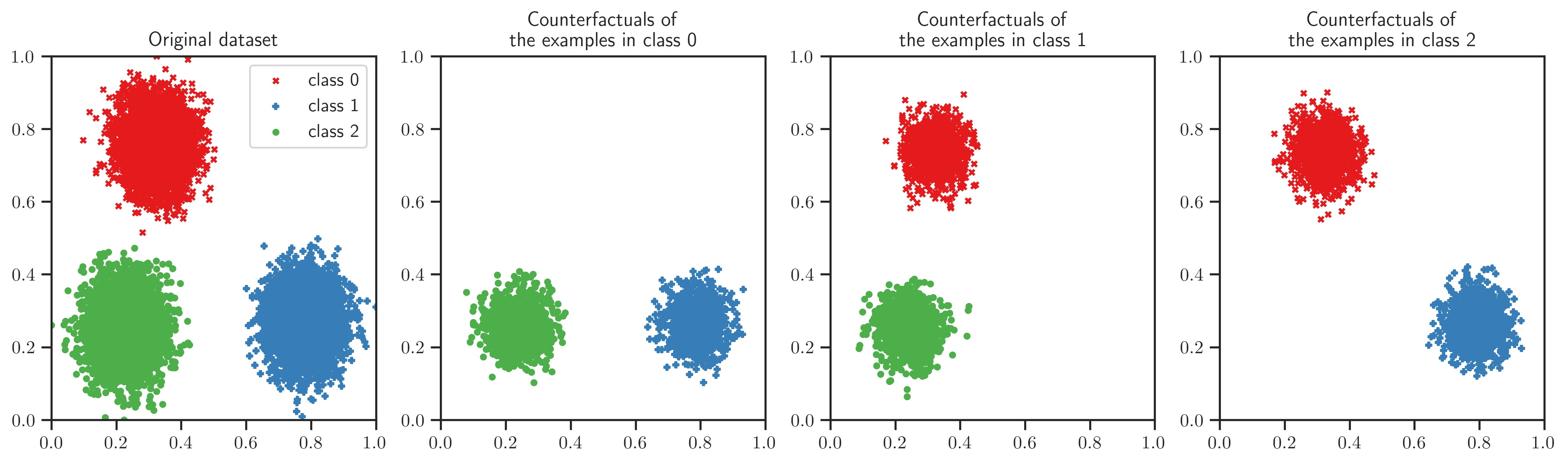}
\caption{Comparison of the examples/counterfactuals distributions for synthetic data. All graphics represent the projection of the examples or counterfacturals on the first two dimensions of the examples space ($\mathbb{R}^8$). The graphic on the left represents the original dataset. The three other graphics represent counterfactuals generated from the examples of the test set belonging to each class. For this three rightmost graphics, the same examples have been used to generate counterfactuals with the two other classes. 
The color/shape of the point represents a class information: the class an example belongs to (graphic on the left) and the class requested for counterfactual generation (3 graphics on the right).  }
\label{fig:expes_synth:validity}
\end{figure}

Our proposal is based on using a cVAE to generate counterfactuals. 
It relies on the capability of a cVAE to actually disentangle the class and its content such that the decoder can be used to generate counterfactual examples by changing the class conditioning.In this experiment, we generate a synthetic dataset of examples in $\mathbb{R}^8$ with three classes. Each class is distributed according to a multidimensional Gaussian distribution (see Figure~\ref{fig:expes_synth:validity} on the left).

A cVAE, i.e. a couple of an encoder $\varphi(\bm{x},c)$ and a decoder $\psi(\bm{z},c)$,  is trained on a set of $10k$ examples. 
The complete code of these experiments is available in supplementary material for the sake of reproducibility.

Figure~\ref{fig:expes_synth:validity} illustrates the capability of a cVAE to generate realistic examples when changing the class that conditions the decoder. 
The figure on the left illustrates the dataset. Each colored group of point corresponds to a class. 
The three figures on the right illustrate datasets that have been generated $\bm{x}'=\psi(\varphi(\bm{x},c),c')$ where $\bm{x}$ is an example from the test that belongs to class $c$ (origin class). Each figure corresponds to one origin class, the colors of the point correspond to the conditioning class ($c'$). 
We observe that all figures look similar. This means that taking $\psi(\varphi(\bm{x},c),c')$ generates an example that looks similar to an example of class $c'$ whatever the origin class of $\bm{x}$. Thus, it demonstrates that cVAE can be used to generate realistic counterfactuals. Moreover, $\bm{x}'$ is a good counterfactual if it is similar to $\bm{x}$. The question is whether $\bm{z}=\varphi(\bm{x},c)$ is a better choice to generate an example $\psi(\bm{z},c')$ than any other example $\psi(\bm{z}',c')$ (which also likely belongs to $c'$). To assess this behavior, we randomly generate 10 latent representations $\bm{z}'=\bm{z}+\bm{\delta}$ for each $\bm{x}$, and compute the Euclidean distance between $\bm{x}'=\psi(\bm{z}',c')$ and $\bm{x}$. 

\begin{figure}[tbp]
\centering
\includegraphics[width=.85\textwidth]{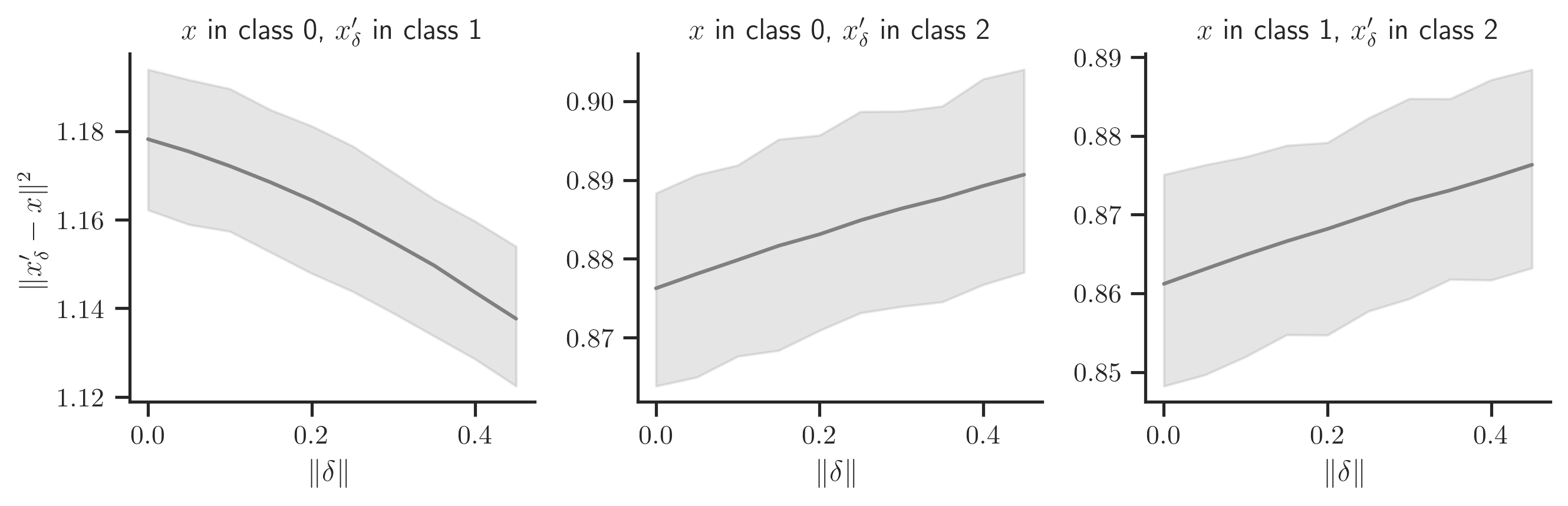}
\caption{Distance between $\bm{x}$, an example of class $c$ to explain, and $\bm{x}'=\psi(\bm{z}+\bm{\delta},c')$, a counterfactual for class $c'$ perturbed in latent space by $\bm{\delta}$. Each graphic illustrates this distance with respect to $\|\bm{\delta}\|$ depending on the class $c$ and $c'$ (on the right: $c=0$ and $c'=1$; in the middle: $c=0$ and $c'=2$; on the left: $c=1$ and $c'=2$). The mean and variance are computed on 10 random $\delta$.}
\label{fig:expes_synth:proximity}
\end{figure}

Figure \ref{fig:expes_synth:proximity} shows three graphs: one per couple of classes (the class of the example to explain and the class requested as counterfactual). 
Each graph illustrates the mean Euclidean distance between $\bm{x}'$ and $\bm{x}$ with respect to $\|\bm{\delta}\|$. 
When $\|\bm{\delta}\|=0$, it uses the latent representation of $\bm{x}$ as input of the cVAE decoder. Intuitively, we expect to have $\psi(\varphi(\bm{x},c),c')$ closer to $\bm{x}$ than $\psi(\varphi(\bm{x},c)+\bm{\delta},c')$ and thus, that the higher $\|\bm{\delta}\|$, the higher the mean distance to the original example. 
The two graphics on the right illustrates the expected behavior. In these cases, the latent representation of the example seems to generate a counterfactual that is the most similar among the examples of the opposite class. 
Nonetheless, we observe that it is not always the case. We can note that the distance decreases when perturbing the latent representation of examples in class 0 and regenerating counterfactual examples in class 1. This may be explained by the proximity between the two classes.

\subsection{Comparison between \VCN and CounterNet}\label{sec:results:cmp_counternet}

This section compares the quality of counterfactuals of \VCN against CounterNet. We conduct evaluations on six binary-classification datasets with various properties: Adult~\cite{kohavi1996uci}, HELOC~\cite{heloc}, OULAD~\cite{kuzilek2017open},  Breast Cancer Wisconsin~\cite{blake1998uci}, Student performance~\cite{cortez2008using} and Titanic~\cite{titanic}. Some of these datasets contain only numerical variables but some others, such as Adult, contain both numerical and categorical variables.
More details about the datasets can be found in the supplementary material of the article.

To evaluate the counterfactual quality, we used the following four metrics that are classical in the literature. 

    \noindent \textbf{Prediction gain:} The prediction gain is given by the difference between the predicted probability of the counterfactual $\bm{x}'$ and the predicted probability of the example $\bm{x}$, according to the counterfactual class~\cite{nemirovsky2020countergan}.
    \[
    \text{Gain} = [f_{\text{pred}}(\bm{x}')]_{y_{i}} - [f_{\text{pred}}(\bm{x})]_{y_{i}} 
    \]
    where $y_{i}$ denotes the predicted class for the counterfactual.
    A higher prediction gain means being more confident in the class change of the counterfactual.

    \noindent \textbf{Validity:} 
    A counterfactual is valid if it achieves to obtain a different predicted class~\cite{MothilalFAT20,mazzine2021framework}. 
    Then:
    \[
      \text{Validity}=\left\{\begin{array}{l}
      0, \text { if } y_{i}=y_{0} \\
      1, \text { if } y_{i} \neq y_{0}
      \end{array}\right.
    \]
    where $y_{i}$ denotes the predicted class for the counterfactual and $y_{0}$ the predicted class for the example to explain.

    \noindent \textbf{Proximity:} The proximity is the $L_{1}$ distance between an example, $\bm{x}$ and its counterfactual, $\bm{x}'$~\cite{MothilalFAT20,Wachter2017CounterfactualEW}. 
    \[
        \text{Proximity} = \left\|\bm{x}'-\bm{x}\right\|_{1} = \| \bm{\delta} \|_{1}
    \]
   A low value indicates fewer changes of features to apply to the original example to obtain the counterfactual. 
   
   \noindent \textbf{Proximity score:} 
   This metric is inspired from Laugel et al.~\cite{laugel} to quantify the distance of a counterfactual to examples of the same predicted class:
   \[
      P_{s}(\bm{x}')=\frac{d\left(\bm{x}', a_{0}\right)}{\frac{1}{\|H\|(\|H\|-1)/2}\sum_{\bm{x}_i,\bm{x}_j\in H}{d\left(\bm{x}_i, \bm{x}_j\right) }
      }
   \]
   where $d\left(\bm{x}', a_{0}\right)$ is the Euclidean distance of the counterfactual to the closest example that has the same predicted class ($a_{0}$) and $H$ is the set of existing examples that have the same predicted class as $\bm{x}'$.
   The insight behind this metric is that the counterfactual should be close to an existing example of the same predicted class relative to the rest of the data. 
   Note that to be evaluated, this metric requires a set of $m$ examples $X\in \mathbb{R}^{m\times p}$.

For each dataset, we choose a random sample of size 25\% for counterfactual generation. Then, we compute the mean and standard deviation of each metric for every selected random sample.

Table~\ref{tab:cf_metrics} provides results of \VCN and CounterNet~\cite{guo2021counternet}.
More information on the reproducibility of CounterNet results is available in Supplementary material.
It is worth noting that Table~\ref{tab:cf_metrics} contains additional results for post-hoc \VCN that will be discussed in Section~\ref{sec:results:posthoc}.

\renewcommand{\arraystretch}{0.9}
\setlength{\tabcolsep}{5pt}
\begin{table}[t!]
\centering
\footnotesize
\caption{Comparison of quality metrics of counterfactuals and predictive accuracy for three methods : \VCN , CounterNet and Post-hoc \VCN. Bold items give the best metric values among the three methods.}\label{tab:cf_metrics}
\begin{tabular}{llll:l}
\hline
\textbf{Datasets}  & \textbf{Metrics} & \multicolumn{3}{c}{\textbf{Methods}} \\ \cmidrule(lr){3-5}
   & & \textbf{\VCN} & \textbf{CounterNet}  & \textbf{Post-hoc \VCN}  \\ \hline 
\textbf{Adult} & \textbf{Validity}  & \textbf{1.0} &  0.99 & 0.84   \\
              & \textbf{Proximity} &  7.71 $\pm$ 2.11 &  \textbf{7.16$\pm$ 2.13} & 7.28$\pm$ 2.23 \\
              & \textbf{Prediction gain} & \textbf{0.76 $\pm$ 0.15} & 0.61$\pm$0.17 & 0.47$\pm$0.35 \\
              & \textbf{Proximity score} & \textbf{0.04 $\pm$ 0.11} & 0.31$\pm$ 0.28 & 0.06$\pm$0.14  \\ \hdashline
              & \textbf{Accuracy} & \textbf{0.83} & \textbf{0.83} & \textbf{0.83} \\ \hline
              
\textbf{OULAD} & \textbf{Validity}   & \textbf{1.0}         & 0.99 &  0.74 \\
              & \textbf{Proximity}  &  11.66$\pm$2.46 & 11.96$\pm$2.40 & \textbf{11.22$\pm$2.54}\\
              & \textbf{Prediction gain} & \textbf{0.93$\pm$0.12} &   0.74$\pm$0.13 & 0.66$\pm$0.44 
              \\ & \textbf{Proximity score} &     \textbf{0.38$\pm$0.18} & 0.46$\pm$0.16 & \textbf{0.38$\pm$0.18}\\\hdashline
              & \textbf{Accuracy} & \textbf{0.93}&  \textbf{0.93} & \textbf{0.93} \\\hline

\textbf{HELOC} & \textbf{Validity}   & \textbf{1.0}  &  0.99 & 0.77 \\
                                        & \textbf{Proximity}  & 5.60$\pm$2.11 & \textbf{4.41$\pm$1.80} & 5.09$\pm$1.71  \\
                                        
                                        & \textbf{Prediction gain}        & \textbf{0.64$\pm$0.13} &  0.56$\pm$0.15 & 0.24$\pm$0.25  \\  
                                        & \textbf{Proximity score} &      \textbf{0.23$\pm$0.21} & 0.49$\pm$0.35 & 0.40$\pm$0.32  \\\hdashline
                                        
                                        & \textbf{Accuracy} &       0.71 & \textbf{0.72} & 0.71  \\\hline
                            
\textbf{Student} & \textbf{Validity}   & 0.96 & \textbf{1.0}  & 0.46                                                    \\
                                        & \textbf{Proximity}  & 19.90$\pm$3.21    & 19.86$\pm$2.78 &  \textbf{19.68 $\pm$3.03}   \\
                                        
                                        & \textbf{Prediction gain}        & \textbf{0.86$\pm$0.27} & 0.76$\pm$0.05 & 0.41$\pm$0.46  \\
                                        & \textbf{Proximity score} &      \textbf{0.70$\pm$0.08} & 0.73$\pm$0.06 & 0.75$\pm$0.08    \\\hdashline
                                        
                                        & \textbf{Accuracy} &       0.90 & \textbf{0.92} & 0.90  \\\hline
                                        
\textbf{Titanic} & \textbf{Validity}   & 0.92 &  \textbf{0.99}  & 0.38                                                     \\
     & \textbf{Proximity}  &  15.43$\pm$3.79 & \textbf{15.15$\pm$4.05} & 15.56$\pm$5.23   \\
     & \textbf{Prediction gain} & \textbf{0.69$\pm$0.31} &  0.66$\pm$0.15 & 0.26$\pm$0.36     \\
     & \textbf{Proximity score} & \textbf{0.71$\pm$0.21} & 0.80$\pm$ 0.16 & 1.21$\pm$0.26    \\\hdashline
     & \textbf{Accuracy} & 0.82 & \textbf{0.83} & 0.82  \\\hline
                                        
\textbf{Breast-} & \textbf{Validity}   & \textbf{1.0} & \textbf{1.0}  & 0.59 \\
\textbf{cancer}  & \textbf{Proximity}  & 5.27$\pm$1.47 & \textbf{1.51$\pm$1.01} & 7.71$\pm$1.67\\
     & \textbf{Prediction gain} & \textbf{0.95 $\pm$ 0.11} & 0.69$\pm$0.15 & 0.60$\pm$0.45 \\
     & \textbf{Proximity score} & \textbf{0.28$\pm$0.03} & 0.72$\pm$0.48 & 0.94$\pm$ 0.07 \\\hdashline
     & \textbf{Accuracy} & \textbf{0.96} & \textbf{0.96} & \textbf{0.96} \\\bottomrule
\end{tabular}    
\end{table}

\textbf{Counterfactual quality:}
\VCN achieves perfect validity for 4 of the 6 datasets, and a lower validity of respectively 4.5\% and 7.5\% for the 2 other datasets (Student and Titanic) compared to CounterNet.

As far as prediction gain and proximity score are concerned, \VCN outperforms CounterNet for all the 6 datasets. The higher the prediction gain, the more confidence one can have in the prediction related to the class change of the counterfactual. At the same time, a low proximity score reflects the achievement of counterfactuals that are close to real examples belonging to the same class as predicted for the counterfactual.

For the last evaluated metric that is  proximity, we observe that \VCN achieves higher values than CounterNet on 5 of the 6 datasets. A larger proximity value means that the counterfactuals are obtained at the cost of larger changes in the input space. 

\textbf{Predictive accuracy}: Both CounterNet and \VCN are self-explainable models, and if the previous results show that \VCN generates better counterfactuals, it can not be at the expense of the prediction accuracy. Thus, Table \ref{tab:cf_metrics} also presents model accuracies.
We observe that \VCN achieved similar performances on 3 of the 6 datasets. 
On the other hand, the accuracies for the other datasets are lower by 0.4\% (HELOC) to 2\% (Student), which shows that our method still performs very well in terms of prediction.

\subsection{Impact of join-training on counterfactual quality} \label{sec:results:posthoc}
\begin{figure}[t!]
    \centering
    \includegraphics[width=\textwidth]{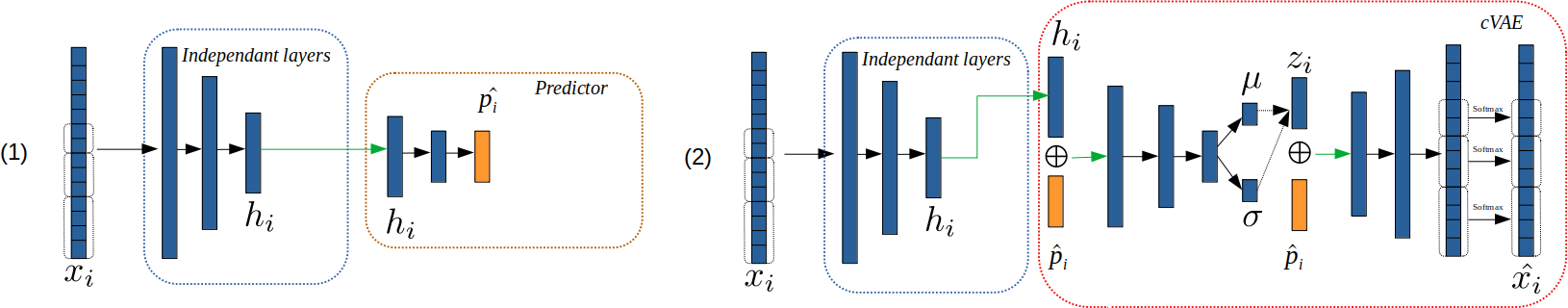}
    \caption{Post-hoc version of \VCN. (1) is the prediction model and (2) is the cVAE model.}
    \label{fig:network_archi_post_hoc}
\end{figure}

We derived our architecture to a post-hoc version (see Figure~\ref{fig:network_archi_post_hoc}). Its training procedure is the following: 
1) we first train a prediction model.
For our comparisons here, it is composed of the concatenation of the shared layers block and the predictor block of \VCN, but in practice it can be any machine learning model that outputs a probability score.
Once the model is trained, we obtain a probability vector $\hat{\bm{p}}_{i}$ 
by forwarding an example $\bm{x}_{i}$ to the model.
2) then we train a cVAE model conditionally to the probability vector~($\hat{\bm{p}}_{i}$) output by the predictor learned during Step 1. 
The cVAE is composed of the same shared layers block than \VCN, but it is not shared with the predictor.

We compare \VCN with its post-hoc version (Post-hoc \VCN) in order to study the impact of join-training on counterfactuals. 
Table~\ref{tab:cf_metrics} provides the results of the post-hoc version compared to those obtained previously with the join-training approach.
We observe a drop of validity for every dataset, which justifies that the join-training approach allows a better alignment of the explanation task with the prediction task. 
We also observe a significant decrease in prediction gain for all datasets, which means less changes between an example to explain and its counterfactual. 
Besides, proximity is higher for 3 datasets (Adult, OULAD, Breast-cancer) and lower by respectively $9\%$, $1\%$ and $0.8\%$ on the remaining datasets (HELOC, Student, Titanic). 
Thus, we can argue that this drop of prediction gain does not benefit to
closer counterfactuals w.r.t. examples to explained.  
In terms of model accuracies, training the prediction model alone leads to comparable results, which indicates that the join-training approach is not at the cost of a lower accuracy.  

\subsection{Qualitative results on MNIST dataset}
We evaluated \VCN on the MNIST dataset\footnote{\url{http://yann.lecun.com/exdb/mnist/}} with metrics suggested in Section~\ref{sec:results:cmp_counternet}. 
This experiment illustrates that \VCN is adaptable to several types of data including image datasets. As CounterNet was not applied to image data in the original paper, we do not offer a comparison with this model here. This avoids a poor adaptation of CounterNet and an unfair comparison. 

\VCN gives a mean validity of $0.99$, a mean prediction gain of $0.98$ and an accuracy of $0.98$. 
The counterfactual quality metrics on MNIST show that \VCN is a good model to generate realistic and valid counterfactuals and, at the same time, to make accurate predictions. 
These results suggest that \VCN has also good capabilities to generate counterfactuals for images, and not only for tabular data.
\begin{figure}[t!]
    \centering
    \includegraphics[width=.9\textwidth]{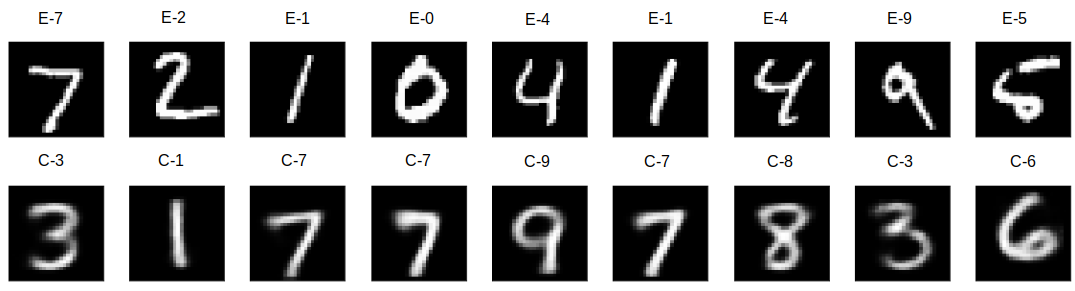}
    \caption{Counterfactuals obtained with \VCN for the MNIST dataset. The top line corresponds to the examples to explain, the bottom to the corresponding counterfactuals. }
    \label{fig:MNIST_counterfactuals}
\end{figure}
Figure~\ref{fig:MNIST_counterfactuals} presents some examples to explain (in the first line) and corresponding counterfactuals generated with \VCN (second line).
Each example to explain and its counterfactual is complemented by the predicted class, for instance \textbf{E-7} means an example predicted class 7 and \textbf{C-3} means a counterfactual predicted class 3. 
We first notice that each counterfactual ``looks like'' an example that matches the predicted class. 
Moreover, we observe that the orientation of the digits is often preserved, for example  \textbf{E-1} is converted in \textbf{C-7} by keeping the orientation of \textbf{E-1}. 
\textbf{C-6} is interesting as it shows that \VCN is able to provide a realistic counterfactual even if the class of the example to explain is not trivial.  

\section{Conclusion}

In this article, we propose \VCN, a new architecture for self-explainable classification based on counterfactuals examples. 
Our architecture generates at the same time the decision and a counterfactual that can be used by the analyst to understand the algorithm decision. 
The first advantage of \VCN is to generate explanations and decisions in a simple feed forward pass of the examples. Contrary to post-hoc counterfactuals explanation, \VCN does not require expensive optimization to generate counterfactuals.

The main contribution of this article is the use of a cVAE as a counterfactual generator in our model. 
The choice of a cVAE yields realistic counterfactuals and it is simple to train jointly with the prediction model.

We extensively evaluated the quality of the counterfactuals and compared them to CounterNet. We conclude that \VCN generates valid and realistic counterfactuals. The \VCN counterfactuals are realistic in the sense that they are close to existing examples of the same predicted class (\VCN has better proximity scores than CounterNet) and also the result of a higher confidence in the class change (\VCN has better prediction gains than CounterNet).  

Finally, \VCN is simple to train because the training procedure is not based on counterfactuals directly, but on the disentanglement of the class and the content of examples by the cVAE. 
It allows proposing a simple optimisation procedure which makes our approach easier to put in practice. 
This is illustrated by its successful application to a dataset of images. 
In addition, it is also assessed in terms of model accuracy. 
Our experiments show that our join-training approach keeps its competitive prediction performance against CounterNet. A future work is to compare \VCN performances against state-of-the-art post-hoc counterfactuals methods and also to include actionability constraints.

\bibliographystyle{plain}
\bibliography{references}

\begin{thebibliography}{10}

\bibitem{NEURIPS2018_3e9f0fc9}
David Alvarez~Melis and Tommi Jaakkola.
\newblock Towards robust interpretability with self-explaining neural networks.
\newblock In {\em Proceedings of the Conference on Advances in Neural
  Information Processing Systems (NIPS)}, pages 7786--7795, 2018.

\bibitem{barr2021counterfactual}
Brian Barr, Matthew~R Harrington, Samuel Sharpe, and C~Bayan Bruss.
\newblock Counterfactual explanations via latent space projection and
  interpolation.
\newblock preprint arXiv:2112.00890, 2021.

\bibitem{blake1998uci}
Catherine Blake.
\newblock {UCI} repository of machine learning databases.
\newblock \url{http://www.ics.uci.edu/mlearn/MLRepository.html}, 1998.

\bibitem{cortez2008using}
Paulo Cortez and Alice Maria~Gon{\c{c}}alves Silva.
\newblock Using data mining to predict secondary school student performance.
\newblock In {\em Proceedings of Annual Future Business Technology Conference},
  pages 5--12, 2008.

\bibitem{mazzine2021framework}
Raphael Mazzine~Barbosa de~Oliveira and David Martens.
\newblock A framework and benchmarking study for counterfactual generating
  methods on tabular data.
\newblock {\em Applied Sciences}, 11(16):7274, 2021.

\bibitem{downs2020cruds}
Michael Downs, Jonathan~L Chu, Yaniv Yacoby, Finale Doshi-Velez, and Weiwei
  Pan.
\newblock Cruds: Counterfactual recourse using disentangled subspaces.
\newblock In {\em ICML Workshop on Human Interpretability in Machine Learning
  (WHI)}, pages 1--23, 2020.

\bibitem{2020Elton}
Daniel~C. Elton.
\newblock Self-explaining {AI} as an alternative to interpretable {AI}.
\newblock In {\em Proceedings of the International Conference on Artificial
  General Intelligence (AGI)}, pages 95--106, 2020.

\bibitem{heloc}
FICO.
\newblock Explainable machine learning challenge.
\newblock
  \url{https://community.fico.com/s/explainable-machine-learning-challenge},
  2018.

\bibitem{guo2021counternet}
Hangzhi Guo, Thanh Nguyen, and Amulya Yadav.
\newblock {CounterNet}: End-to-end training of counterfactual aware
  predictions.
\newblock In {\em ICML Workshop on Algorithmic Recourse}, 2021.

\bibitem{johnetal2019disentangled}
Vineet John, Lili Mou, Hareesh Bahuleyan, and Olga Vechtomova.
\newblock Disentangled representation learning for non-parallel text style
  transfer.
\newblock In {\em Proceedings of the Annual Meeting of the Association for
  Computational Linguistics (ACL)}, pages 424--434, 2019.

\bibitem{titanic}
Kaggle.
\newblock Titanic -- machine learning from disaster.
\newblock \url{https://www.kaggle.com/c/titanic/overview}, 2018.

\bibitem{VAE_original}
Diederik~P. Kingma and Max Welling.
\newblock Auto-encoding variational {Bayes}.
\newblock In {\em Proceedings of the International Conference on Learning
  Representations (ICLR)}, 2014.

\bibitem{kingma2014semi}
Durk~P Kingma, Shakir Mohamed, Danilo Jimenez~Rezende, and Max Welling.
\newblock Semi-supervised learning with deep generative models.
\newblock In {\em Proceedings of International Conference on neural information
  processing systems (NIPS)}, pages 3581--3589, 2014.

\bibitem{kohavi1996uci}
R~Kohavi and B~Becker.
\newblock {UCI} machine learning repository: Adult data set, 1996.

\bibitem{kuzilek2017open}
Jakub Kuzilek, Martin Hlosta, and Zdenek Zdrahal.
\newblock Open university learning analytics dataset.
\newblock {\em Scientific data}, 4:170171, 2017.

\bibitem{laugel}
Thibault Laugel, Marie-Jeanne Lesot, Christophe Marsala, Xavier Renard, and
  Marcin Detyniecki.
\newblock The dangers of post-hoc interpretability: Unjustified counterfactual
  explanations.
\newblock In {\em Proceedings of the International Joint Conference on
  Artificial Intelligence (IJCAI)}, pages 2801--2807, 2019.

\bibitem{molnar2022}
Christoph Molnar.
\newblock {\em Interpretable Machine Learning}.
\newblock C. Molnar, 2nd edition, 2022.

\bibitem{MothilalFAT20}
Ramaravind~K. Mothilal, Amit Sharma, and Chenhao Tan.
\newblock Explaining machine learning classifiers through diverse
  counterfactual explanations.
\newblock In {\em Proceedings of the Conference on Fairness, Accountability,
  and Transparency (FAT)}, pages 607--617, 2020.

\bibitem{Nangi2021CounterfactualsTC}
Sharmila~Reddy Nangi, Niyati Chhaya, Sopan Khosla, Nikhil Kaushik, and Harshit
  Nyati.
\newblock Counterfactuals to control latent disentangled text representations
  for style transfer.
\newblock In {\em Proceedings of the Annual Meeting of the Association for
  Computational Linguistics (ACL)}, pages 40--48, 2021.

\bibitem{nemirovsky2020countergan}
Daniel Nemirovsky, Nicolas Thiebaut, Ye~Xu, and Abhishek Gupta.
\newblock {CounteRGAN}: Generating realistic counterfactuals with residual
  generative adversarial nets.
\newblock preprint arXiv:2009.05199, 2020.

\bibitem{PawelczykWWW20}
Martin Pawelczyk, Klaus Broelemann, and Gjergji Kasneci.
\newblock Learning model-agnostic counterfactual explanations for tabular data.
\newblock In {\em Proceedings of The Web Conference (WWW'20)}, pages
  3126--3132, 2020.

\bibitem{2019rudin}
Cynthia Rudin.
\newblock Stop explaining black box machine learning models for high stakes
  decisions and use interpretable models instead.
\newblock {\em Nature Machine Intelligence}, 1:206--215, 2019.

\bibitem{RusselFAT19}
Chris Russell.
\newblock Efficient search for diverse coherent explanations.
\newblock In {\em Proceedings of the Conference on Fairness, Accountability,
  and Transparency (FAT)}, New York, NY, USA, 2019. Association for Computing
  Machinery.

\bibitem{CVAE_original}
Kihyuk Sohn, Honglak Lee, and Xinchen Yan.
\newblock Learning structured output representation using deep conditional
  generative models.
\newblock In {\em Proceedings of the Conference on Advances in Neural
  Information Processing Systems (NIPS)}, pages 3483--3491, 2015.

\bibitem{UstunFAT19}
Berk Ustun, Alexander Spangher, and Yang Liu.
\newblock Actionable recourse in linear classification.
\newblock In {\em Proceedings of the Conference on Fairness, Accountability,
  and Transparency (FAT)}, pages 10--19, 2019.

\bibitem{VanLooverenECML21}
Arnaud Van~Looveren and Janis Klaise.
\newblock Interpretable counterfactual explanations guided by prototypes.
\newblock In {\em Proceedings of the European Conference on Machine Learning
  and Knowledge Discovery in Databases (ECML/PKDD)}, pages 650--665, 2021.

\bibitem{Wachter2017CounterfactualEW}
Sandra Wachter, Brent~Daniel Mittelstadt, and Chris Russell.
\newblock Counterfactual explanations without opening the black box: Automated
  decisions and the {GDPR}.
\newblock {\em Harvard Journal of Law and Technology}, 31(2):841--887, 2018.

\end{thebibliography}

\appendix

\section{Dataset Details}
\subsection{Tabular Data} 
We used 6 binary-classification datasets for the comparison of \VCN with CounterNet. 
Table \ref{tab:dataset} describe the size of each dataset as well as the number of categorical and continuous variables. 

\subsection{MNIST Dataset}
MNIST is composed of $60000$ examples in train and $10000$ examples in test.  
We used all of the test set for counterfactual generation.

\begin{table*}[t]
\centering
\small
\caption{\label{tab:dataset}Datasets details}
\begin{tabular}{llcc}
\hline \text { Dataset } & \text { Size } & \text { Continuous } & \text { Categorical } \\
\hline \text { Adult } & 32,561 & 2 & 6 \\
\text { Student } & 649 & 2 & 14 \\
\text { Titanic } & 891 & 2 & 24 \\
\text { HELOC } & 10,459 & 21 & 2 \\
\text { OULAD } & 32,593 & 23 & 8 \\
\text { Breast Cancer } & 569 & 30 & 0 \\
\hline
\end{tabular}
\end{table*}

\section{Implementation Details}

\subsection{\VCN Details} 
We choose a binary cross entropy loss as the reconstruction error for the cVAE part and a cross entropy loss for the predictor part. 
We apply a grid search to tune hyperparameters that are specific for each datasets, these values are report in Table \ref{tab:hyperparameter_tabular}. 
We used an Adam optimizer for every dataset. 
Table \ref{tab:architecture_tabular} describe the model architecture for every dataset.
It contains the dimensions for each block of \VCN.
For example, with the adult dataset, the shared encoding transforms an example $\bm{x}_i$ of size \textbf{29} into a vector $\bm{h}_i$ of size \textbf{15}.
Then the encoder of the cVAE part takes a concatenation of $\bm{h}_i$ and the prediction vector $\hat{\bm{p}}_{i}$  to produce a vector of size \textbf{16}. 
This vector is transform by the encoder to a lower representation of size \textbf{8} and finally a vector of size \textbf{5}.

\subsection{Post-hoc \VCN For Tabular Data} 
For the post-hoc version of \VCN, the architecture is the same as detailed in Table \ref{tab:architecture_tabular}.
Nonetheless, hyperparameters for training change.
Table \ref{tab:hyperparameter_tabular_post_hoc_vae} and \ref{tab:hyperparameter_tabular_post_hoc_predictor} gives training hyperparameters for the cVAE model and prediction model respectively.

\begin{table*}[t]
\centering
\small
\caption{\label{tab:hyperparameter_tabular}Hyperparameters details for \VCN}
\begin{tabular}{llccccc}
\hline \textbf{Dataset} & \textbf{$\lambda_1$} & \textbf{$\lambda_2$} & \textbf{  $\lambda_3$ } & \textbf{Learning Rate} & \textbf{Epochs}  & \textbf{Batch-size} \\
\hline \text { Adult } & 1 & 1 & 0.001 & 0.001 & 250 & 128 \\
\text { Student } & 1 & 0.1 & 0.01 & 0.001 & 50 & 30  \\
\text { Titanic } & 1 & 1 & 0.001 & 0.001 & 100 & 30\\
\text { HELOC } & 0.1 & 1 & 0.01 & 0.001 & 200 & 64 \\
\text { OULAD } & 1 & 1 & 0.01 & 0.001 & 40 & 128 \\
\text { Breast Cancer } & 1 & 0.1 & 0.001 & 0.001 & 100 & 30 \\
\text { MNIST } & 1 & 8 & 0.2 & 0.001 & 100 & 30 \\
\hline
\end{tabular}
\end{table*}

\begin{table*}[t]
\centering
\small
\caption{\label{tab:architecture_tabular}Architecture details for \VCN}
\begin{tabular}{llccc}
\hline \textbf{Dataset} & \textbf{Shared Encoding Dims} & \textbf{Encoding Dims} & \textbf{Predictor Dims} & \textbf{CF Generator Dims}\\
\hline \text { Adult } & [29,15] & [16,8,5] & [6,8,15,29] & [15,15,1] \\
\text { Student } & [85,50] & [51,20,10] & [11,20,50,85] & [50,50,1] \\
\text { Titanic } & [57,20] & [21,10,5] & [6,10,20,57] & [20,20,1] \\
\text { HELOC } & [35,15] &  [16,8,5] & [6,8,15,35] & [15,15,1] \\
\text { OULAD } &  [127,200] & [201,100,10] & [11,100,200,127] & [200,200,1] \\
\text { Breast Cancer }  & [30,15] & [16,8,5] & [6,8,15,30] & [15,15,1] \\
\text { MNIST }  & [784,400,40] & [50,400,20] & [30,400,784] & [40,10] \\
\hline
\end{tabular}
\end{table*}

\begin{table*}[t]
\centering
\small
\caption{\label{tab:hyperparameter_tabular_post_hoc_vae}Hyperparameters details for post-hoc \VCN on tabular data for the cVAE part}
\begin{tabular}{llcccc}
\hline \textbf{Dataset} & \textbf{$\lambda_2$} & \textbf{  $\lambda_3$ } & \textbf{Learning Rate} & \textbf{Epochs}  & \textbf{Batch-size} \\
\hline \text { Adult } & 1 & 0.001 & 0.001 & 10 & 128 \\
\text { Student } & 1 & 0.01 & 0.001 & 50 & 30\\
\text { Titanic } & 1 & 0.001 & 0.001 & 20 & 30\\
\text { HELOC } & 0.1 & 0.01 & 0.001 & 100 &  64\\
\text { OULAD } &  0.1 & 0.01 & 0.001 & 40 & 128\\
\text { Breast Cancer } & 1 & 0.001 & 0.001 & 30 &  30  \\
\hline
\end{tabular}
\end{table*}

\begin{table*}[t]
\centering
\small
\caption{\label{tab:hyperparameter_tabular_post_hoc_predictor}Hyperparameters details for post-hoc \VCN on tabular data for the prediction part}
\begin{tabular}{llcc}
\hline \textbf{Dataset} & \textbf{Learning Rate} & \textbf{Epochs}  & \textbf{Batch-size} \\
\hline \text { Adult } & 0.001 & 50 & 128 \\
\text { Student } & 0.001 & 50 & 30  \\
\text { Titanic } & 0.001 & 80 & 30\\
\text { HELOC } & 0.001 & 100 & 64\\
\text { OULAD } &  0.001 & 150 & 128 \\
\text { Breast Cancer } & 0.001 & 10 &  30  \\
\hline
\end{tabular}
\end{table*}

\subsection{Additional Information on CounterNet Reproducibility}
We used the CounterNet code that was available at this link \url{https://github.com/bkghz-orange-blue/CounterNet}.
For fair comparison, we have computed counterfactuals from available trained models and used the same data processing for \VCN.

\end{document}